\documentclass{article} % For LaTeX2e
\usepackage{nips15submit_e,times}
\usepackage{hyperref}
\usepackage{url}

\usepackage{graphicx}
\usepackage{caption}
\usepackage{subcaption}
\usepackage{MnSymbol}
\usepackage{float}

\title{Efficient ResNets: Residual Network Design\thanks{Project done as part of the course on Deep Learning at NYU Tandon School of Engineering.}}

\author{%
Aditya Thakur\thanks{All authors contributed equally to this work.} \quad Harish Chauhan\footnotemark[2] \quad Nikunj Gupta\footnotemark[2] \\
New York University\\
\texttt{\{at4932, hpc4473, nikunj.gupta\}@nyu.edu} 
}

\nipsfinalcopy % Uncomment for camera-ready version

\begin{document}

\maketitle 

\begin{abstract}
ResNets (or Residual Networks) are one of the most commonly used models for image classification tasks. In this project, we designed and trained our own modified ResNet model for CIFAR-10 image classification. In particular, we aimed at maximizing the test accuracy on the CIFAR-10 benchmark while keeping the size of our ResNet model under the specified fix budget of 5 million trainable parameters. Model size, typically measured as the number of trainable parameters, is important when models need to be stored on devices with limited storage capacity (e.g. IoT/edge devices). In this article, we present our residual network design which has less than 5 million parameters. We show that our ResNet achieves a test accuracy of 96.04\% on CIFAR-10 which is much higher than ResNet18 (which has greater than 11 million trainable paramters) when equipped with a number of training strategies and suitable ResNet hyperparameters. Models and code are available at \href{https://github.com/Nikunj-Gupta/pytorch-cifar}{https://github.com/Nikunj-Gupta/Efficient\_ResNets}.

\end{abstract}

\section{Introduction} 
ResNet is one of the most popular deep neural networks used in computer vision-related problems. The general trend has been to make deeper and more complicated networks to achieve higher accuracy. However, these advances to improve accuracy are not necessarily making networks more efficient concerning size and speed. In many real-world applications such as embedded, robotics, self-driving car and augmented reality, the recognition tasks need to be carried out in a timely fashion on a computationally limited platform. Hence, there is an increasing need for making memory-efficient models that offer competitive performance. Related works like MobileNet[7]and WideNet[8] have attempted to address this problem. In this article, we describe our attempt to find a memory-efficient configuration for the ResNet-18 model. The subsequent sections provide an overview of our approach, the parameters we tweaked, the optimization methods we tried to finally reach an optimum configuration in which we have less than 50\% trainable parameters and better accuracy in comparison to the original ResNet-18 model.

\section{Methodology: ResNet Hyperparameters}

\subsection{Residual Layers (N) and Residual blocks in Residual Layer i ($B_i$)} 
\begin{figure}[H]
\centering
\includegraphics[width=0.75\linewidth]{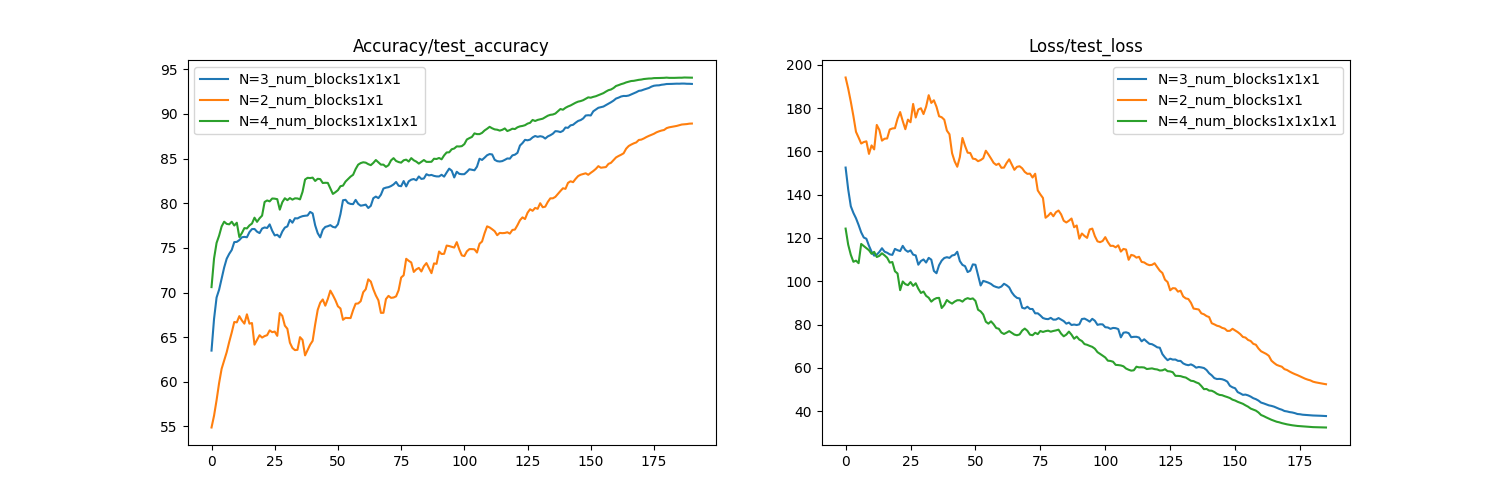}
\caption{Varying the number of residual layers and blocks}
\label{plot:num_layers_blocks}
\end{figure}

To meet the criteria of using less than 5 million trainable parameters, the maximum number of residual layers we could use is 4. We performed experiments on several configurations including different number of residual layers and blocks (Figure \ref{plot:num_layers_blocks} shows the learning curves corresponding to a small/elementary subset of configurations to illustrate the intuitive conclusion that deeper networks perform better on a complex task like image classification). With residual blocks, inputs can forward propagate faster through the residual connections across layers. By configuring different numbers of channels and residual blocks in the module, we can create different ResNet models. 

\subsection{Number of Channels in Residual Layer i ($C_i$)} 

Let us observe the changes in the input shape on changing the number of channels in ResNet. The resolution decreases and the number of channels increases up until the point where a global average pooling layer aggregates all features.

\begin{center}
\begin{tabular}{ c c }
 Sequential output shape:  & torch.Size([1, n, 32, 32]) \\ 
 Sequential output shape:  & torch.Size([1, 2*n, 16, 16]) \\  
 Sequential output shape:  & torch.Size([1, 4*n, 8, 8]) \\    
 Sequential output shape:  & torch.Size([1, 8*n, 4, 4]) \\ 
 AvgPool2d output shape &   torch.Size([1, 8*n, 1, 1]) \\ 
 Flatten output shape &  torch.Size([1, 512]) \\ 
 Linear output shape  &   torch.Size([1, 10]) \\ 

\end{tabular}
\end{center}

\begin{figure}[H]
\centering
\includegraphics[width=0.75\linewidth]{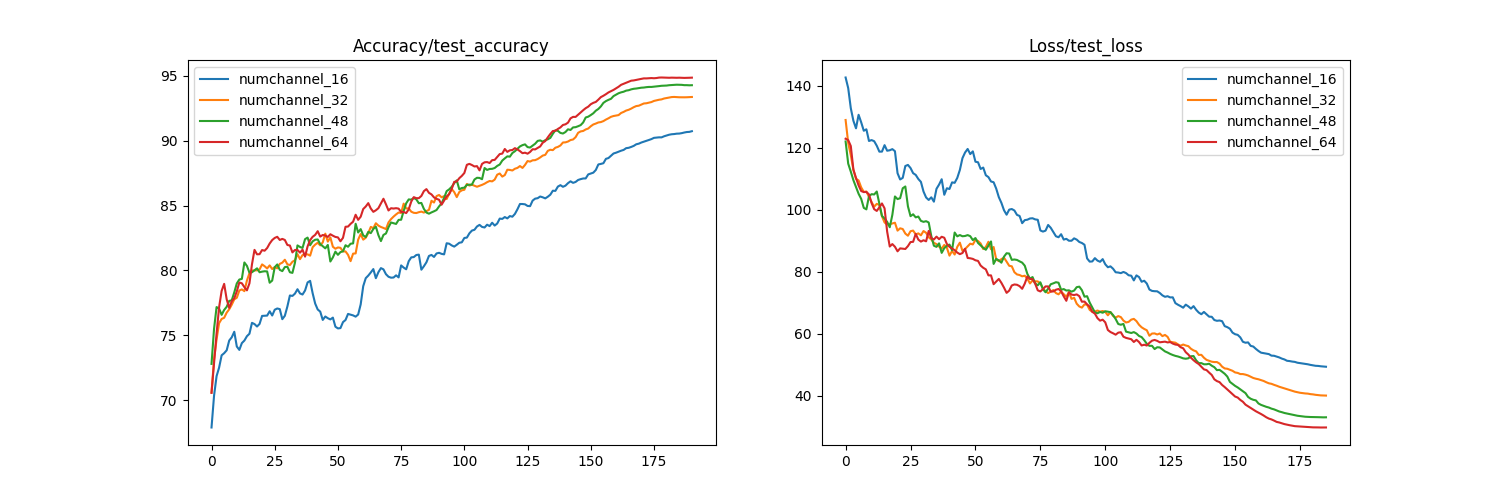}
\caption{Varying the number of channels in residual layers }
\label{plot:numchannels}
\end{figure}

We test for n=\{16, 32, 48, 64\}. The results are show in Figure \ref{plot:numchannels}

\subsection{Convolutional kernel sizes in Residual Layer i ($F_i$)} 

\begin{figure}[H]
\centering
\includegraphics[width=0.75\linewidth]{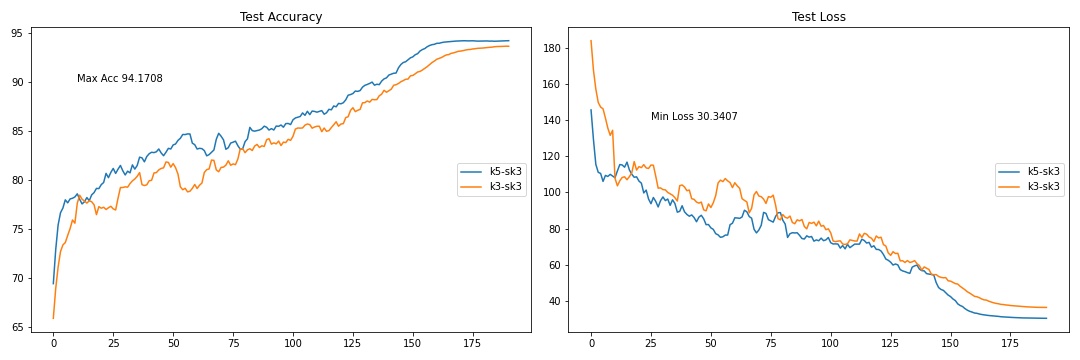}
\caption{Varying Convolutional Kernel size}
\label{plot:conv_kernel}
\end{figure}

Convolutional neural networks help us extract underlying low level features from local view points in an image. The number of parameters grows quadratically with kernel size. Consequently, large convolution kernels are not cost efficient, which further intensifies when we want a large number of channels. We tried a few kernel sizes, and observed what works best. A convolution kernel of size greater than 3 only satisfies the constraint of 5M parameters for 3 or less layers. As an example, figure \ref{plot:conv_kernel} shows the learning curve for convolutional kernel sizes ranging from 1 to 3 keeping the other parameters (shown below) as constant: \\ 

Block Size: [2,1,1] \\
Number of Channels: [64,128,256] \\ 
Conv. Kernel Size: 5 \\ 
Skip Kernel Size: 3 \\ 

\subsection{Ki: Skip connection kernel size in Residual Layer i} In our analysis we tried varying skip connection kernel size for a 3 layered network. Within the constraint of 5M parameters, we could experiment with a kernel size as large as 9 and evidently the kernel size 9 did yield best accuracy.

\begin{figure}[H]
\centering
\includegraphics[width=0.75\linewidth]{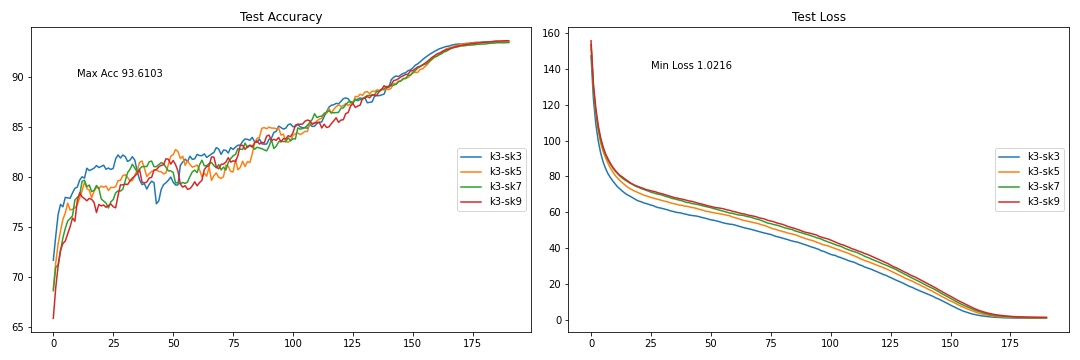}
\caption{Varying Convolutional Kernel size and Skip Kernel sizes}
\label{plot:skip_kernel}
\end{figure}

\subsection{Average pool kernel size (P)} Average Pooling is a pooling operation that calculates the average value for patches of a feature map, and uses it to create a downsampled (pooled) feature map. It is usually used after a convolutional layer. It adds a small amount of translation invariance - meaning translating the image by a small amount does not significantly affect the values of most pooled outputs. It extracts features more smoothly than Max Pooling, whereas max pooling extracts more pronounced features like edges. 

For our usecase, because we restrict ourselves with fixed strides, no changes in image resolutions within a residual layer and an upper bound of 5M parameters, the kernel size for the average pool layer just depends on the number of layers we have in our ResNet architecture. As the resolution of the image decreases by a factor of 2 in every residual layer, we can use the following formula to determine the value for average pool kernel (P) 

\[P = \frac{32}{2^{N-1}}\] 

Here, N is the number of residual layers we have in our architecture.

\section{Methodology: Training Strategies} 

\subsection{Data preparation strategies} 

There are a few important changes we made while creating the PyTorch datasets: 

\paragraph{Use test set for validation:} Instead of setting aside a fraction (e.g. 10\%) of the data from the training set for validation, we'll simply use the test set as our validation set. This just gives a little more data to train with. In general, once you have picked the best model architecture \& hyperparameters using a fixed validation set, it is a good idea to retrain the same model on the entire dataset just to give it a small final boost in performance.
    
\paragraph{Channel-wise data normalization:} We will normalize the image tensors by subtracting the mean and dividing by the standard deviation across each channel. As a result, the mean of the data across each channel is 0, and standard deviation is 1. Normalizing the data prevents the values from any one channel from disproportionately affecting the losses and gradients while training, simply by having a higher or wider range of values that others.

\paragraph{Randomized data augmentations:} We will apply randomly chosen transformations while loading images from the training dataset. Specifically, we will pad each image by 4 pixels, and then take a random crop of size 32 x 32 pixels, and then flip the image horizontally with a 50\% probability. Since the transformation will be applied randomly and dynamically each time a particular image is loaded, the model sees slightly different images in each epoch of training, which allows it generalize better.

\begin{figure}[H]
\centering
\includegraphics[width=0.75\linewidth]{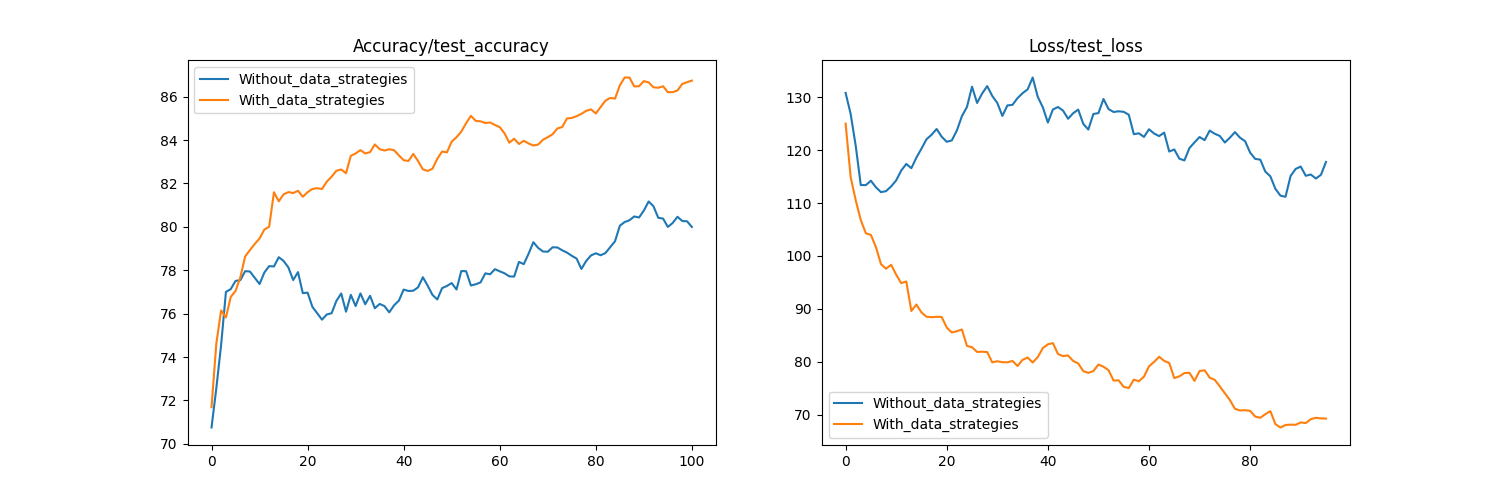}
\caption{Data Augmentation and Data Normalization techniques help in learning better models.}
\end{figure}

\subsection{Optimizers, Learning rates (LR) and LR schedulers} 
\begin{figure}[H]
\centering
\includegraphics[width=0.75\linewidth]{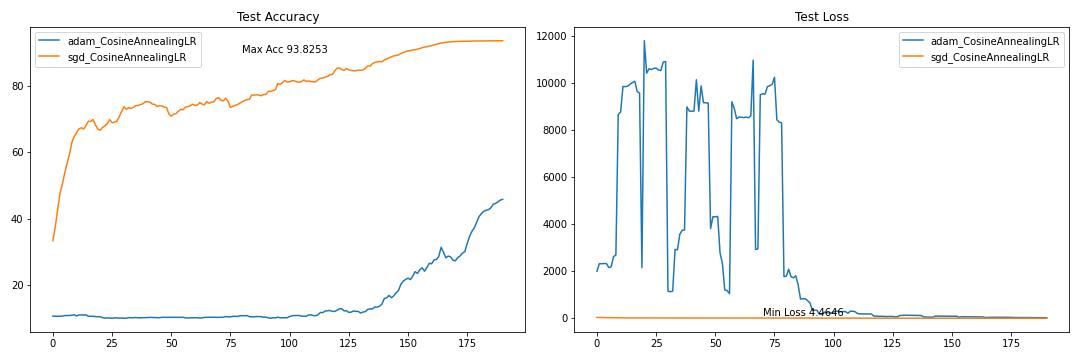}
\caption{SGD Optimizer vs. ADAM}
\label{plot:sgd_adam}
\end{figure}

\paragraph{Optimizers:} Optimizers are used to minimize a loss function to improve the accuracy of the model as well as aid to the fast learning. There are various optimization algorithms present, we chose Stochastic Gradient Descent and Adam(Adaptive Moment Estimation) to run our experiments on CIFAR-10.\\
We observed for 200 epochs SGD performed better. ADAM performs better with slow learning rates, therefore the learning convergence for ADAM is comparatively slow. Hence we selected SGD optimizer for our final model. The results from our experiments can be seen in figure \ref{plot:sgd_adam}.

\paragraph{Learning Rate:} We analyzed the impact of varying learning rate, momentum and batch size. For a fixed learning rate we varied momentum and batch sizes and we found out that the best learning rate and batch size combination is when \textbf{"Learning Rate = 0.1 and Batch Size = 128"} and the best learning rate momentum combination is when \textbf{Learning Rate = 0.2 and Momentum = 0.8}
\begin{figure}[H]
     \centering
     \begin{subfigure}[b]{0.3\textwidth}
         \centering
         \includegraphics[width=\textwidth]{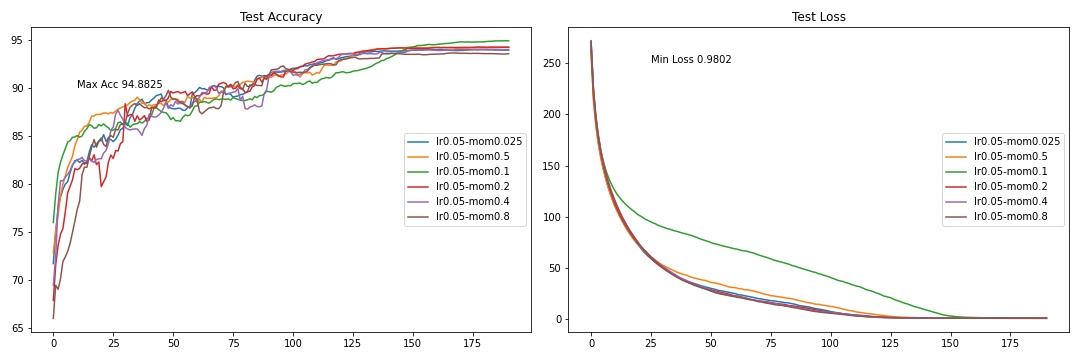}
         \caption{Lr = 0.05}
        %  \label{fig:y equals x}
     \end{subfigure}
     \hfill
     \begin{subfigure}[b]{0.3\textwidth}
         \centering
         \includegraphics[width=\textwidth]{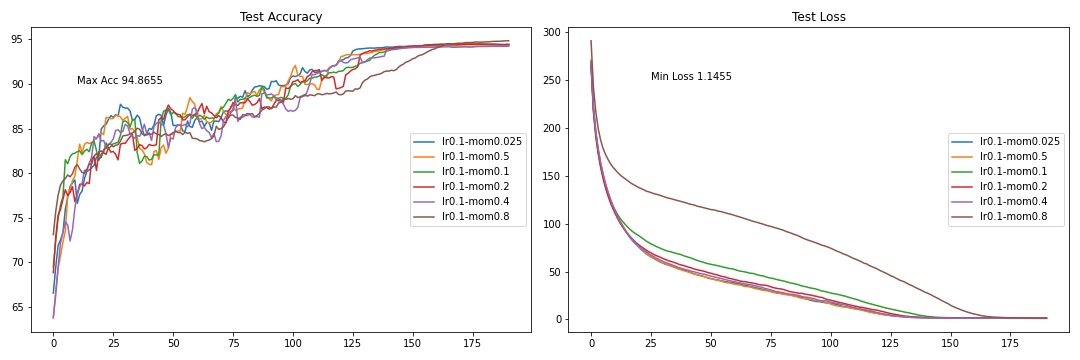}
         \caption{Lr = 0.1}
        %  \label{fig:y equals x}
     \end{subfigure}
     \hfill
     \begin{subfigure}[b]{0.3\textwidth}
         \centering
         \includegraphics[width=\textwidth]{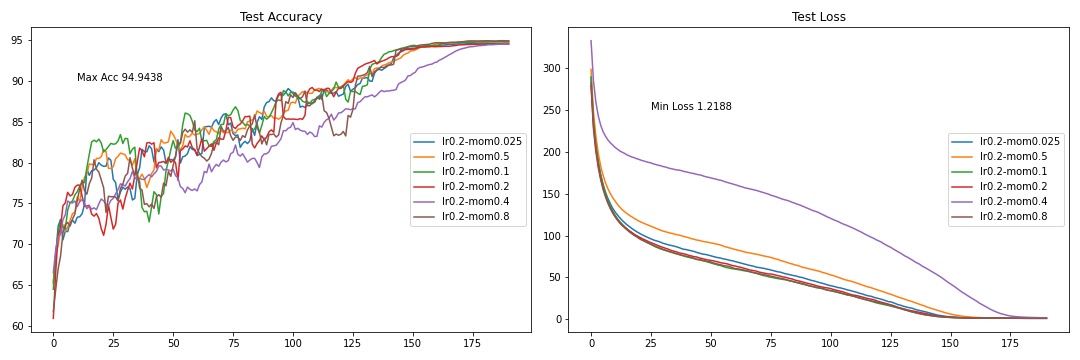}
         \caption{Lr = 0.2}
        %  \label{fig:y equals x}
     \end{subfigure}
     \begin{subfigure}[b]{0.3\textwidth}
         \centering
         \includegraphics[width=\textwidth]{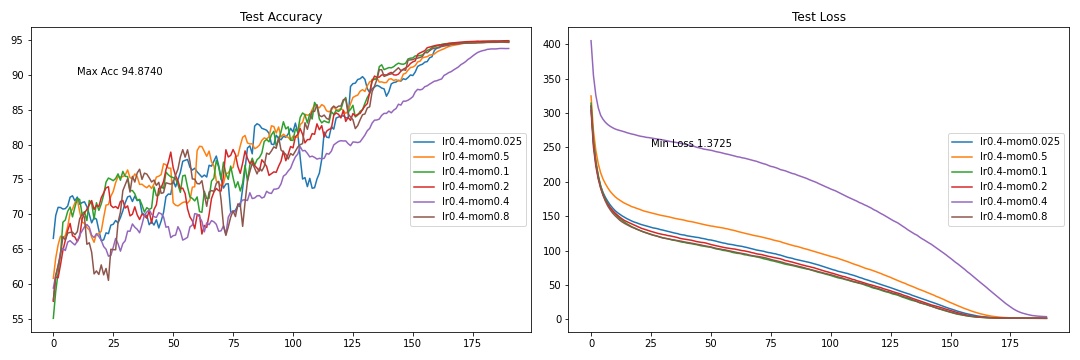}
         \caption{Lr = 0.4}
        %  \label{fig:y equals x}
     \end{subfigure}
     \hfill
     \begin{subfigure}[b]{0.3\textwidth}
         \centering
         \includegraphics[width=\textwidth]{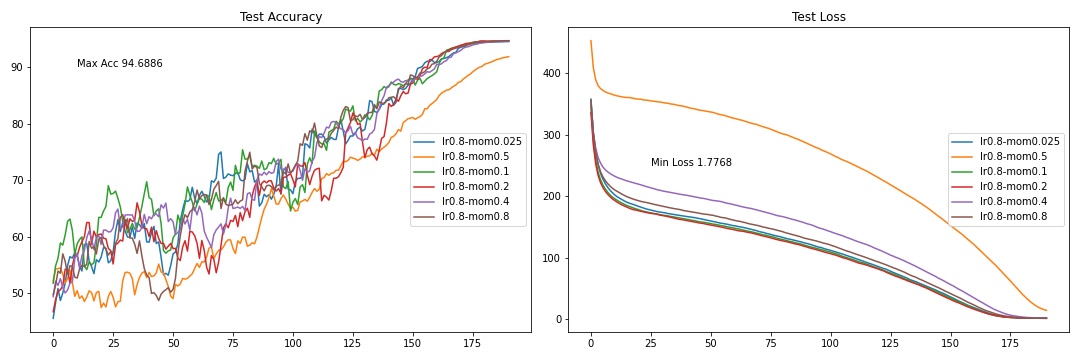}
         \caption{Lr = 0.8}
        %  \label{fig:y equals x}
     \end{subfigure}
     \hfill
        \caption{Learning Rate vs Momentum}
        \label{fig:Lr-Momentum}
\end{figure}

\begin{figure}[H]
     \centering
     \begin{subfigure}[b]{0.3\textwidth}
         \centering
         \includegraphics[width=\textwidth]{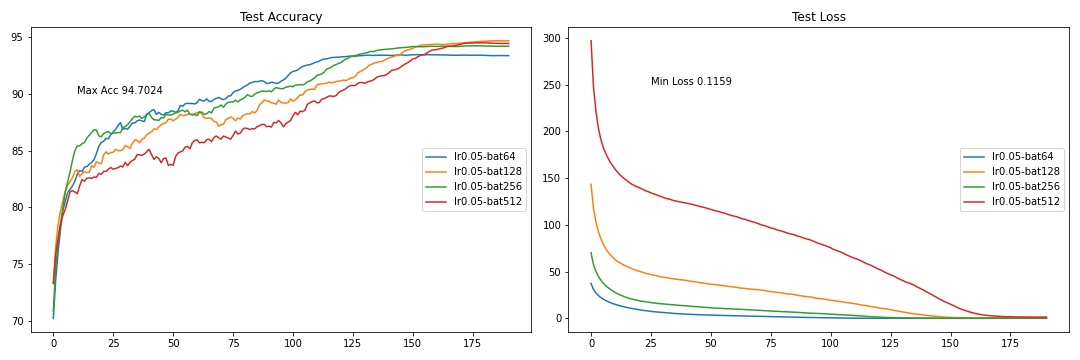}
         \caption{Lr = 0.05}
        %  \label{fig:y equals x}
     \end{subfigure}
     \hfill
     \begin{subfigure}[b]{0.3\textwidth}
         \centering
         \includegraphics[width=\textwidth]{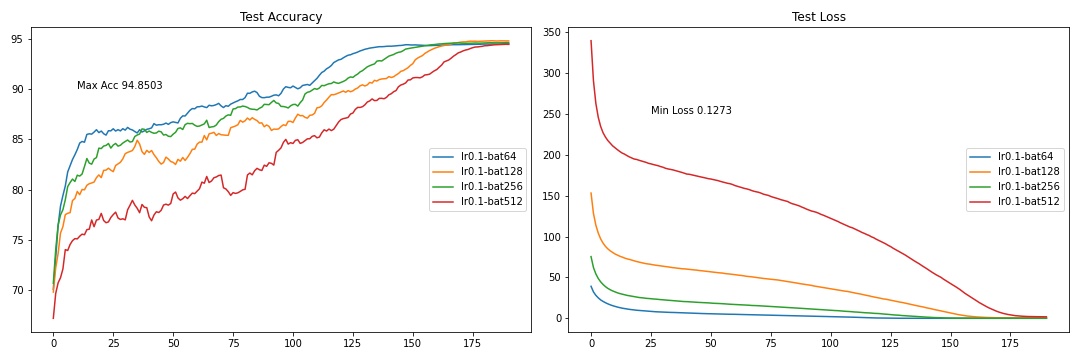}
         \caption{Lr = 0.1}
        %  \label{fig:y equals x}
     \end{subfigure}
     \hfill
     \begin{subfigure}[b]{0.3\textwidth}
         \centering
         \includegraphics[width=\textwidth]{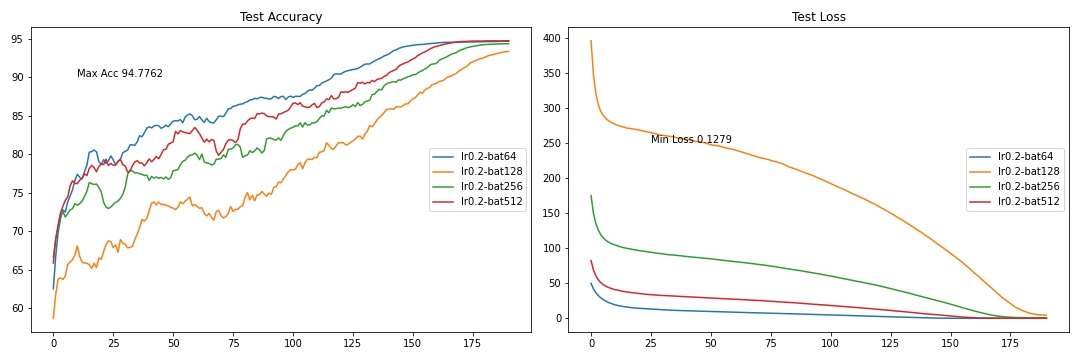}
         \caption{Lr = 0.2}
        %  \label{fig:y equals x}
     \end{subfigure}
     \begin{subfigure}[b]{0.3\textwidth}
         \centering
         \includegraphics[width=\textwidth]{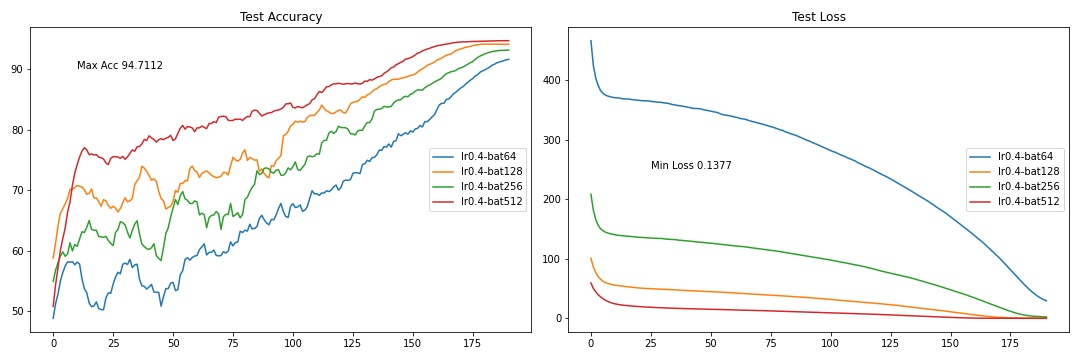}
         \caption{Lr = 0.4}
        %  \label{fig:y equals x}
     \end{subfigure}
     \hfill
     \begin{subfigure}[b]{0.3\textwidth}
         \centering
         \includegraphics[width=\textwidth]{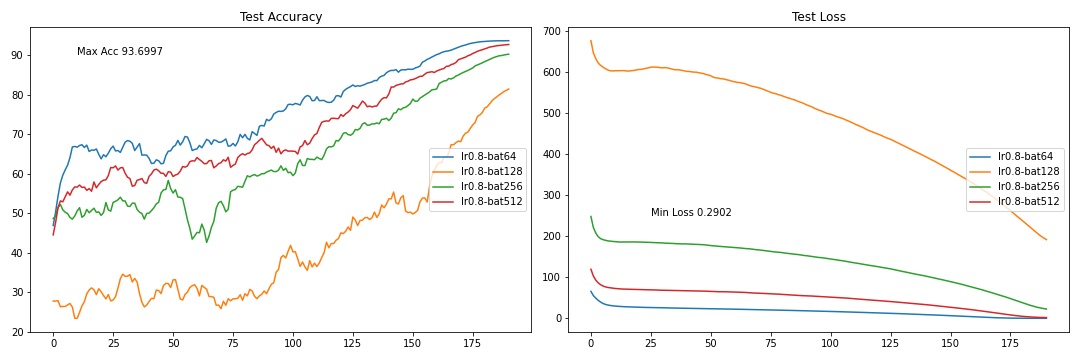}
         \caption{Lr = 0.8}
        %  \label{fig:y equals x}
     \end{subfigure}
     \hfill
        \caption{Learning Rate vs Batch Size}
        \label{fig:Lr-batchsize}
\end{figure}

\paragraph{Learning Rate Schedulers:}
Learning rate scheduler provides a framework which alters the learning rate between epochs during the training. We ran our experiments with the following 12 learning rate schedulers provided in the PyTorch [CosineAnnealingLR,          
LambdaLR,                
MultiplicativeLR,           
StepLR,                     
MultiStepLR,                
ExponentialLR,              
CyclicLR,                   
CyclicLR2,                  
CyclicLR3,                  
OneCycleLR,                 
OneCycleLR2,                
CosineAnnealingWarmRestarts ]

\begin{figure}[H]
\centering
\includegraphics[width=0.75\linewidth]{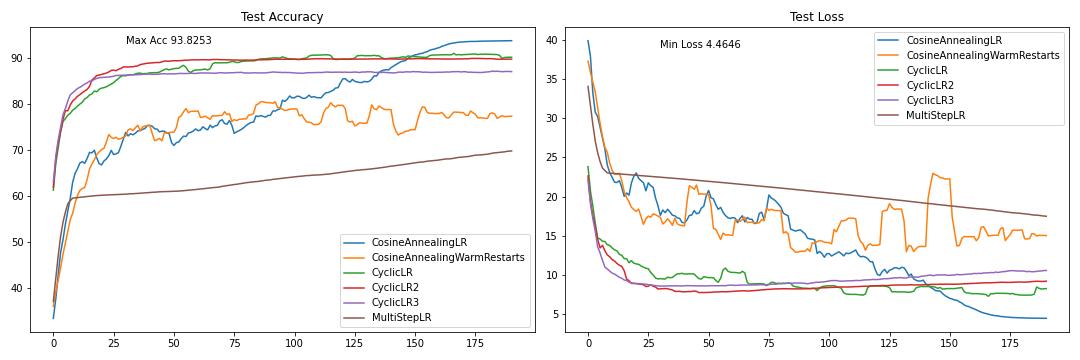}
\caption{SGD Optimizer with different LR Schedulers}
\label{plot:sgd_lr}
\end{figure}

In the figure \ref{plot:sgd_lr} there are 6 out of 12 learning rate schedulers with which we did our experiments for test accuracy and test loss. Empirically we found that CosineAnnealingLR has the best performance on CIFAR-10 dataset with the given 200 epochs.

\subsection{Gradient Clipping} 
\paragraph{Gradient clipping:} To avoid the issue of exploding gradients, we are using gradient clipping technique. In gradient clipping, if a gradient becomes too large, it is rescaled so that it remains small.
More precisely, if $||g|| \geq c$, then $g\leftmapsto c . \frac{g}{||g||}$, where $c$ is a hyperparameter, g is the gradient, and $||g||$ is the norm of g [4].
Gradient clipping bounds the gradient vector $g$ has norm at most $c$. 
With gradient clipping, the descent step size is restricted and even if there is a very large rise in the loss landscape the parameters stay in the good region.

\begin{figure}[H]
\centering
\includegraphics[width=0.75\linewidth]{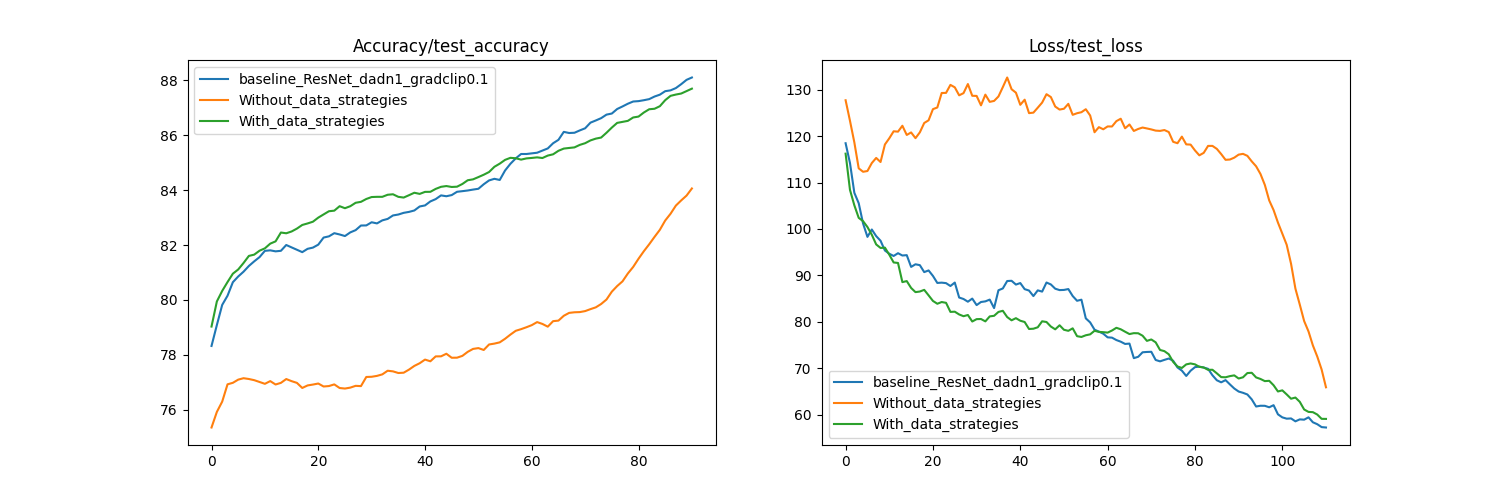}
\caption{Gradient clipping helped preventing exploding gradients for our model.}
\end{figure}

\subsection{Lookahead} 
Lookahead[5] is an optimization method, that is orthogonal to already existing approaches like SGD, Adam etc. It first updates the “fast weights” (elastic weight which stores temporary knowledge and spontaneously decays towards zero) k times using any standard optimizer in its inner loop before updating the “slow weights” (which stores long-term knowledge) once in the direction of the final fast weights . This reduces variance and Lookahead becomes less sensitive to suboptimal hyperparameters and therefore lessens the need for extensive hyperparameter tuning. By using Lookahead with inner optimizers such as SGD or Adam, faster convergence can be achieved across different deep learning tasks with minimal computational overhead. The graph below confirms that convergence happens faster with lookahead without much loss of overall efficiency.
\begin{figure}[!h]
\centering
\includegraphics[width=0.75\linewidth]{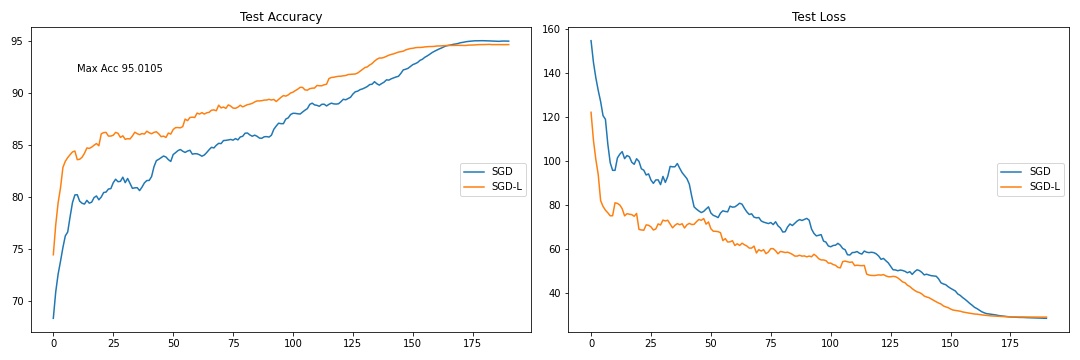}
\caption{SGD with Lookahead helps in faster convergence}
\end{figure}
\subsection{Network Weight Initialization} 
There are a number of important, and sometimes subtle, choices that need to be made when building and training a neural network. You have to decide which loss function to use, how many layers to have, what stride and kernel size to use for each convolution layer, which optimization algorithm is best suited for the network, etc. With so many things that need to be decided, the choice of initial weights may, at first glance, seem like just another relatively minor pre-training detail, but weight initialization can actually have a profound impact on convergence rate,  vanishing gradients and final quality of a network. \\

As part of our analysis, we tried number of weight initialization approaches which are natively available in Pytorch framework. The following graph shows a comparison between He, Xavier and Normal weights initialization.

\begin{figure}[H]
\centering
\includegraphics[width=0.75\linewidth]{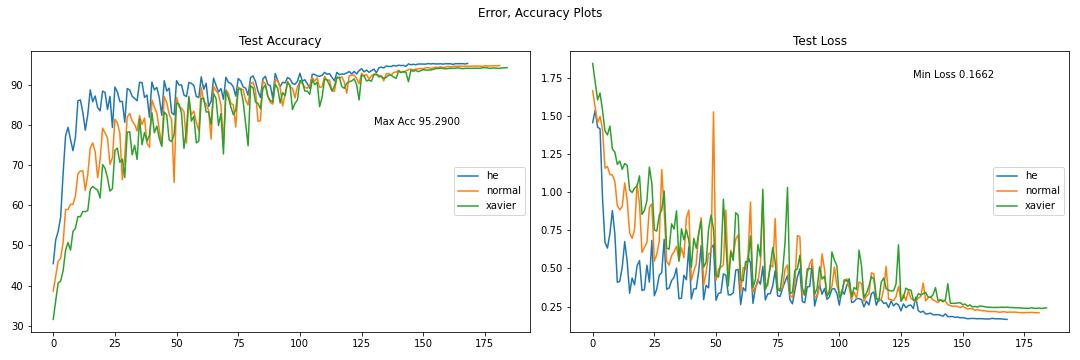}
\caption{Weight Initialization}
\end{figure}

\subsection{Regularization using Batch Normalization and Dropout} 

To justify the need for \textbf{batch normalization}, consider the following scenario:

Suppose we use a batch of low-dimensional data that has a low value of mean and standard deviation to train the model. Now, the second batch we use has a different mean and standard deviation. Now, suppose the following few batches again have a low mean and standard deviation. Consequently, the model will have to change its weights by a large value to accommodate the second batch. Also, now for the next few batches that have a lower mean and standard deviation, the model will again perform poorly, simply because it has adapted for the significantly deviating scenario. In other words, due to differences in batches of training data, model optimization will oscillate quite heavily, which is undesirable, because it slows down the training process. This can be handles by batch normalization, i.e., normalizing the inputs to each layer to a learnt representation close to ($\mu$ = 0.0, $\sigma$ = 1.0). [1]

% Batch Normalization allows us to use much higher learning rates and be less careful about initialization, and in some cases eliminates the need for Dropout. Applied to a stateof-the-art image classification model, Batch Normalization achieves the same accuracy with 14 times fewer training steps, and beats the original model by a significant margin. Using an ensemble of batch-normalized networks, we improve upon the best published result on ImageNet classification: reaching 4.82\% top-5 test error, exceeding the accuracy of human raters. 

Overfitting is a serious problem in deep neural nets with a large number of parameters. \textbf{Dropout} is a technique for addressing this problem. The key idea is to randomly drop units (along with their connections) from the neural network during training. This prevents units from co-adapting too much. [2]

\begin{figure}[H]
\centering
\includegraphics[width=0.75\linewidth]{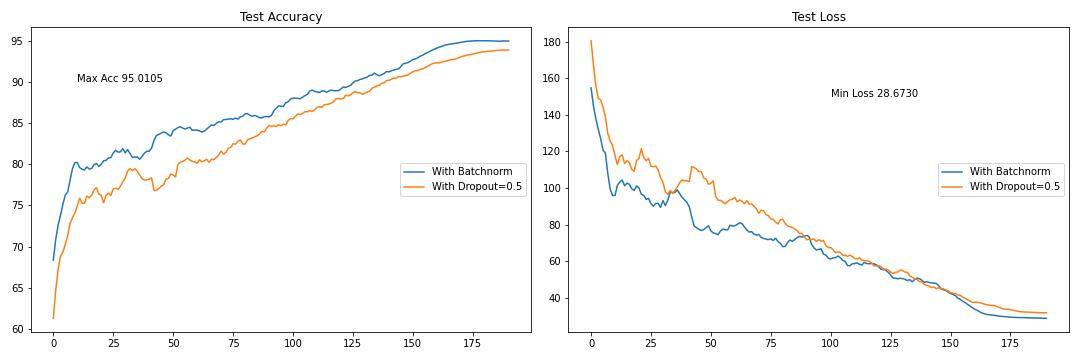}
\caption{Effect of batch normalization and dropout on Test Accuracy and Test Loss}
\label{plot:batchnorm_dropout}
\end{figure}

In figure \ref{plot:batchnorm_dropout} we are comparing results with dropout, where we observe that, on adding dropout layer on top of Batch normalization increases the test loss compared to scenario when only batch normalization is used.

\subsection{Squeezing and Excitation} 

\begin{figure}[H]
\centering
\includegraphics[width=0.5\linewidth]{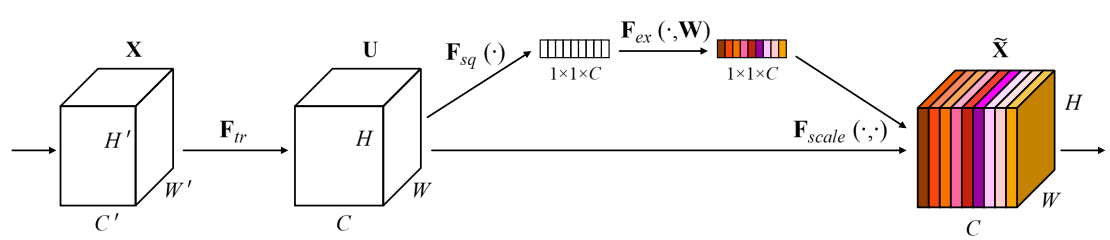}
\caption{A Squeeze-and-Excitation block.}
\label{fig:se_block}
\end{figure}

Squeeze-and-excitation block is a new architecture. It helps in recalibrating channel-wise feature responses by modeling the interdependency between the channels [3]. It consists of two successive operations: \textit{squeeze} (to compress the feature maps along the spatial dimension and extracting the mean of the feature maps for each channel) and \textit{excitation} (to model the correlation between the channels, and then generate a weight for each channel). Figure \ref{fig:se_block}) illustrates a squeeze-and-excitation block. The output of the squeeze-and-excitation block is generated by multiplying the block's input feature maps with the output weights of the excitation operation. 

\begin{figure}[H]
\centering
\includegraphics[width=0.75\linewidth]{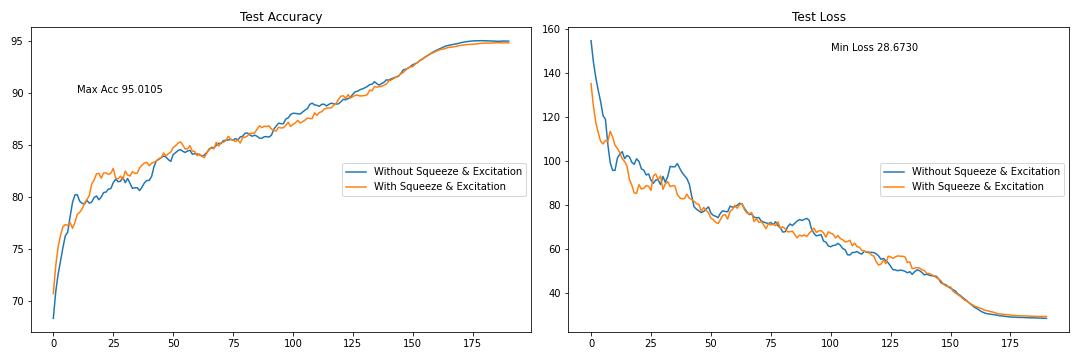}
\caption{Effect of Squeeze and Excitation on Test Accuracy and Test Loss }
\label{plot:2111_se}
\end{figure}

As seen in figure \ref{plot:2111_se}, in our experiments with Squeeze-and-excitation, test accuracy shows subtle improvement with added Squeeze-and-excitation block.
\section{Results and Discussion} 

\begin{figure}[htp]
\centering
\includegraphics[width=0.75\linewidth]{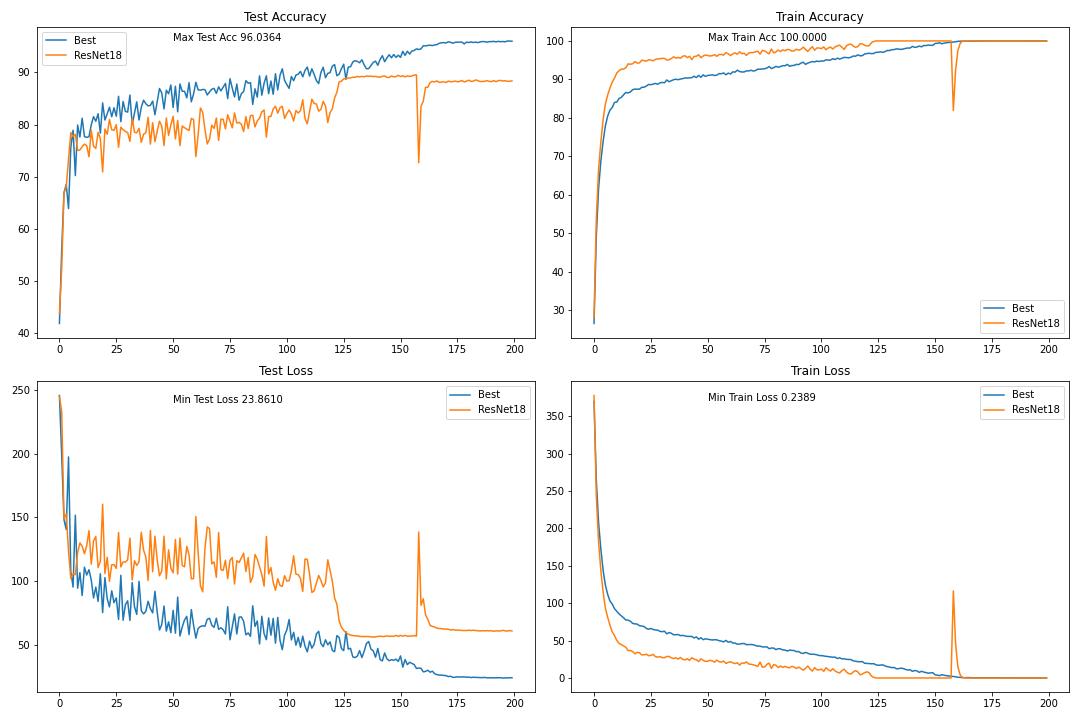}
\caption{Learning curves compare the performance of our Residual Network Design and ResNet18 (hyperparameters listed in table \ref{tab:final})}
\label{plot:best}
\end{figure}

In this section, we compare the learning curves of ResNet18 and our best model achieved after manipulating several hyperparameters upon ResNet18 making sure we use lesser than 5 million trainable parameters. As shown in Figure \ref{plot:best}, our model achieves a final test accuracy of close to 96\%, whereas ResNet18, trained without any of the training strategies mentioned in previous section (including data augmentation and data normalization), converges with a test accuracy of almost 89\%. 

Table \ref{tab:final} shows the final configuration of hyperparameters that were used for our Residual Network Design and for Resnet18 used fro training.

\begin{table}[h]
\begin{tabular}{ |p{5cm}||p{3cm}||p{3cm}| }
 \hline
 \multicolumn{3}{|c|}{\textbf{Hyperparameters}} \\
 \hline
 \textbf{Parameter} & \textbf{Our Model} & \textbf{ResNet18}\\
 \hline
 number of residual layers & 3 & 4\\
 number of residual blocks & [4, 4, 3]& [2, 2, 2, 2] \\
 convolutional kernel sizes & [3, 3, 3]  & [3, 3, 3, 3]\\
 shortcut kernel sizes & [1, 1, 1] & [1, 1, 1, 1] \\
 number of channels   & [64, 128, 256]  & [64, 128, 256, 512]   \\
 average pool kernel size   & 8   & 4 \\
 batch normalization & True & True \\
 dropout & 0 & 0\\
 squeeze and excitation & True  & False\\
 gradient clip & 0.1 & None\\
 data augmentation   & True  & False  \\
 data normalization &   True  & False\\
 lookahead & True & False\\
 optimizer & SGD & SGD\\
 learning rate (lr) & 0.1 & 0.1\\
 lr scheduler & CosineAnnealingLR & CosineAnnealingLR\\
 weight decay & 0.0005 & 0.0005 \\
 batch size & 128 & 128\\
 number of workers   & 16  & 16   \\
\hline
 \textbf{Total number of Parameters} & \textbf{4,697,742} & \textbf{11,173,962} \\
\hline
\end{tabular}
\caption{The final configuration for our Residual Network Design that achieves an accuracy of 96.04\%}
\label{tab:final}
\end{table}

\section{Conclusion}

Using our model, we achieve a test accuracy of greater than 96\% by using less than half the number of parameters compared to that used by ResNet18, which achieves a test accuracy close to 90\%.

\subsubsection*{Acknowledgments}

The model we built is ResNet-18 [6] and certain modifications in its hyperparameters followed by training the model with several new and useful strategies. Reference code for this article can be found at \href{https://github.com/Nikunj-Gupta/Efficient_ResNets}{https://github.com/Nikunj-Gupta/pytorch-cifar}. We built upon the well-written base code by Kuang Liu available at \href{https://github.com/kuangliu/pytorch-cifar}{https://github.com/kuangliu/pytorch-cifar}. The reference code for squeeze-and-excitation block is available at \href{https://github.com/osmr/imgclsmob/blob/68335927ba27f2356093b985bada0bc3989836b1/pytorch/pytorchcv/models/common.py#L731}{https://github.com/osmr/imgclsmob}. 

\newpage 
 
 \section*{References}
 \small{

[1] Ioffe, S., \& Szegedy, C. (2015, June). Batch normalization: Accelerating deep network 
training by reducing internal covariate shift. In International conference 
on machine learning (pp. 448-456). PMLR. 

[2] Srivastava, N., Hinton, G., Krizhevsky, A., Sutskever, I., \& Salakhutdinov, R. (2014). Dropout: a simple way to prevent neural networks from overfitting. The journal of machine learning research, 15(1), 1929-1958.

[3] Hu, J., Shen, L., \& Sun, G. (2018). Squeeze-and-excitation networks. In Proceedings of the IEEE conference on computer vision and pattern recognition (pp. 7132-7141). 

[4] Pascanu, R., Mikolov, T., \& Bengio, Y. (2013, May). On the difficulty of training recurrent neural networks. In International conference on machine learning (pp. 1310-1318). PMLR.

[5] Zhang, M., Lucas, J., Ba, J., \& Hinton, G. E. (2019). Lookahead optimizer: k steps forward, 1 step back. Advances in Neural Information Processing Systems, 32.

[6] He, K., Zhang, X., Ren, S., \& Sun, J. (2016). Deep residual learning for image recognition. In Proceedings of the IEEE conference on computer vision and pattern recognition (pp. 770-778). 

[7] Howard, A. G., Zhu, M., Chen, B., Kalenichenko, D., Wang, W., Weyand, T., ... \& Adam, H. (2017). Mobilenets: Efficient convolutional neural networks for mobile vision applications. arXiv preprint arXiv:1704.04861.

[8] Xue, F., Shi, Z., Wei, F., Lou, Y., Liu, Y., \& You, Y. (2021). Go wider instead of deeper. arXiv preprint arXiv:2107.11817.
}
\end{document}